\documentclass[letterpaper, 10pt, conference]{ieeeconf} 

\IEEEoverridecommandlockouts 

\UseRawInputEncoding  

\usepackage{amsmath,amsfonts}
\usepackage{algorithmic}
\usepackage{algorithm}
\usepackage{array}
\usepackage[caption=false,font=normalsize,labelfont=sf,textfont=sf]{subfig}
\usepackage{textcomp}
\usepackage{stfloats}
\usepackage{url}
\usepackage{verbatim}
\usepackage{graphicx}
\usepackage{cite}
\usepackage[colorlinks=true,
            linkcolor=red,
            pagebackref=true]{hyperref}  
\usepackage{multirow}
\usepackage{graphicx}
\usepackage{makecell}

\title{\LARGE \bf A Superalignment Framework in Autonomous Driving with Large Language Models}

\author{Xiangrui Kong$^{1,2}$, Thomas Braunl$^{1}$, Marco Fahmi$^{2}$, and Yue Wang$^{2,3}$ 
\thanks{*This work was supported in part by Australian Postgraduate Research Intern (APR.Intern) under reference number APR-2384, and INT-1256.}%
\thanks{$^{1}$The authors are with the Department of Electrical, Electronic and Computer Engineering, University of Western Australia, Crawley, WA 6009, Australia. E-mail:
{\tt\small xiangrui.kong@research.uwa.edu.au, thomas.braunl@uwa.edu.au}}%
\thanks{$^{2}$The authors are with the Department of Transport and Main Roads, Queensland Government, Brisbane, QLD 4000, Australia. E-mail: {\tt\small Marco.Fahmi@chde.qld.gov.au}}%
\thanks{$^{3}$The authors are with the Center for Data Science, Queensland University of Technology, Brisbane, QLD 4000, Australia. E-mail: {\tt\small y355.wang@hdr.qut.edu.au}}%
}

\begin{document}

\maketitle

\begin{abstract}
Over the last year, significant advancements have been made in the realms of large language models (LLMs) and multi-modal large language models (MLLMs), particularly in their application to autonomous driving. These models have showcased remarkable abilities in processing and interacting with complex information. In autonomous driving, LLMs and MLLMs are extensively used, requiring access to sensitive vehicle data such as precise locations, images,  and road conditions. This data is transmitted to an LLM-based inference cloud for advanced analysis. However, concerns arise regarding data security, as the protection against data and privacy breaches primarily depends on the LLM's inherent security measures, without additional scrutiny or evaluation of the LLM's inference outputs. Despite its importance, the security aspect of LLMs in autonomous driving remains underexplored.
Addressing this gap, our research introduces a novel security framework for autonomous vehicles, utilizing a multi-agent LLM approach. This framework is designed to safeguard sensitive information associated with autonomous vehicles from potential leaks, while also ensuring that LLM outputs adhere to driving regulations and align with human values. It includes mechanisms to filter out irrelevant queries and verify the safety and reliability of LLM outputs. 
Utilizing this framework, we evaluated the security, privacy, and cost aspects of eleven large language model-driven autonomous driving cues. Additionally, we performed QA tests on these driving prompts, which successfully demonstrated the framework's efficacy.
\end{abstract}

\section{introduction}
Large Language Models (LLMs) have gained significant attention recently, showing remarkable potential in emulating human-like intelligence \cite{llmadcui2023survey}.
A core challenge for aligning future superhuman AI systems (superalignment) is that humans will need to supervise AI systems much smarter than them \cite{burns2023weaktostrong}. 
The transformer-based network structure, mainly Generative Pre-trained Transformer (GPT) such as GPT-3 \cite{floridi2020gpt3}, and Llama2 \cite{llama2touvron2023llama}, transfers the complexity of the data to the complexity of the network, and demonstrates powerful text reasoning and understanding capabilities. 
More and more autonomous systems are using LLMs as the interaction portal between humans and machines, including robots \cite{driess2023palm} and autonomous vehicles \cite{ezmilellmDriveLLM}.
At present, the research on the interaction between LLM and unmanned systems is still in its infancy. 
Since LLMs need to perform inference on higher-power computing devices, the current mobile architecture cannot provide stable electrical power and computing power to support offline inference of LLMs.
A common framework is to use LLMs in the cloud for inference and obtain the inference results of LLM through cloud service calls.


\begin{figure}[!t]
	\centering
	\captionsetup[subfloat]{labelfont=scriptsize,textfont=scriptsize}
	\subfloat[Insecure LLM-AD framework]{\includegraphics{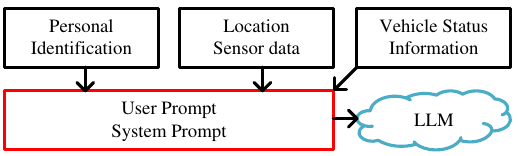}%
		\label{fig:framework1}}
	\hfil
	\subfloat[Proposed LLM-AD framework]{\includegraphics{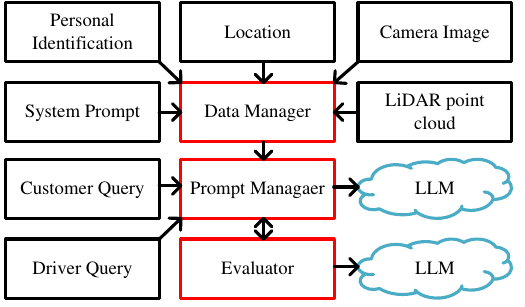}%
		\label{fig:framework2}}
	\caption{LLM Safety-as-a-service autonomous driving framework}
	\label{fig:framework}
\end{figure}

These LLM-driven autonomous agents has the following risks.
First of all, decision-making reasoning for autonomous agent requires uploading a large amount of sensitive information such as image data, precise location, and personal information, which poses the risk of data leakage. 
Secondly, LLMs also face inherent challenges, such as being prone to subtle biases, arithmetic inaccuracies, and the risk of hallucinations.
When LLM-driven unmanned systems interact with the environment, these built-in risks will be reflected in the real-world environment, leading to unknown consequences.
Finally, the inference output results of LLM may not conform to the numerical values in specific situations, thereby violating local laws, regulations or customs, leading to a reduction in people's trust in LLMs.

The main contributions of this paper are summarized as follows:
\begin{itemize}
	\item Propose a secure interaction framework for LLM, which serves as a guardrail between vehicles and cloud LLM, effectively censoring the data interacting with cloud-based LLM.
	\item  We analyzed eleven autonomous driving methods based on large language models, including driving safety, token usage, privacy, and the alignment of human values.
	\item Utilizing our framework, we assessed the effectiveness of driving prompts within a segment of the nuScenes-QA dataset and compared the varying outcomes between the gpt-35-turbo and llama2-70b LLM backbones.
\end{itemize}


\section{Related work}
\subsection{LLMs in Autonomous Driving}
The knowledge is included in the LLMs not only for language tasks, but also for making goal-driven decisions in interactive environments \cite{huang2022language}.
LanguageMPC \cite{mpcsha2023languagempc} employs LLMs to forecast vehicular dynamics, utilizing a bird's-eye view (BEV) to comprehend interactive situations or roundabout scenarios, alongside the consideration of the vehicles' current status.
The Agent-Driver \cite{agentdrivermao2023language} method develops an LLM-driven framework capable of processing a variety of driving information, including images, point clouds, driving rules, and maps, which allows the LLM to access and interpret this diverse data through function calls, utilizing a chain-of-thought approach for comprehensive analysis.
The DriveLLM \cite{drivellmcui2023} method integrates rule-based driving methods with LLMs, implementing the LLM for campus driving scenarios, and demonstrates high real-time performance within a stable network, evidenced by the efficient token processing time in GPT-3.5.

Currently, there exists a notable gap in the security research concerning the application of pre-trained large AI models in autonomous driving.
Self-driving cars are at risk of potentially harmful or malicious activity when interacting with cloud systems \cite{BOUCHOUIA2023100586}.
This process entails detecting and countering attempts to jam or disrupt communication signals, discerning and addressing false or misleading information, and responding to efforts to hack or compromise the vehicle's systems \cite{thing2016autonomous}.
The survey \cite{misbehavior10015746} referenced identifies various common Non-IP-based attacks on autonomous vehicles, such as position falsification \cite{so2018integrating}, dissemination of false information \cite{singh2020misbehavior}, Sybil attacks \cite{kamel2019misbehavior}, and privacy issues \cite{uprety2021privacy}.
With the growing incorporation of LLMs in autonomous driving applications, the range of these attack methods is expected to expand.

\subsection{Privacy and Alignment in LLMs}

As both the model and data size increase, generative LLMs show a promising ability to understand and are capable of integrating classification tasks into their generative pipelines \cite{JMLR:v21:20-074}.
The safety issues related to LLMs have recently garnered widespread attention \cite{tjondronegoro2022responsible}. 
Although Differential Privacy (DP) \cite{dwork2014algorithmic} provides a theoretical worst-case privacy guarantee for safeguarded data, current privacy mechanisms considerably diminish the utility of LLMs, making many existing approaches impractical.

In the realm of LLMs, recent research has identified three safety areas of concern: prompt injection, data breaches, and model hallucinations. 
The phenomenon of prompt injection emerges as a significant security risk, wherein specifically crafted inputs are utilized to manipulate or exploit the natural language processing capabilities of AI systems. 
Moreover, LLMs are susceptible to inadvertent data breaches, where sensitive information may be leaked through model outputs, often attributed to the incorporation of confidential datasets during the training phase \cite{namer2023cost}. 
Additionally, a critical issue identified in these models is their tendency towards hallucination, where they generate erroneous or illogical information, often with a false sense of confidence, due to limitations in their predictive text generation algorithms \cite{martino2023knowledge}. 
These findings underscore the need for enhanced security measures and algorithmic refinements in the development and deployment of LLMs to mitigate these risks.

In the burgeoning field of artificial intelligence, the alignment of LLMs with human and organizational values presents a critical area of research, necessitating a multifaceted approach to ensure ethical and effective AI deployment \cite{wolf2023fundamental}. 
In current research on LLMs, alignment of output text is primarily influenced through two methods.
Firstly, the training data of the LLM significantly impacts its alignment, shaping the nature of the generated content \cite{NEURIPS2022_b1efde53}. 
Secondly, LLM service providers offer optional API alignment services, designed to filter out content that starkly deviates from predefined norms or standards \cite{governAI}. 
Additionally, LLM customers often customize alignment requirements to suit their specific needs, typically employing simpler methods such as Retrieval-Augmented Generation (RAG) \cite{lewis2020retrieval} or tailored prompting techniques. 

\section{Method}

In order to model the behavior of LLM and alignment tasks, we follow the theoretical approach called Behavior Expectation Bounds (BEB) \cite{wolf2023fundamental}. 
The behavior scoring functions are defined along a vertical axis $\displaystyle B$ as $\displaystyle B:{\textstyle \Sigma ^{*}\rightarrow [ -1,1]}$. 
These functions evaluate a text string from an alphabet $\displaystyle \Sigma $, assessing how the behavior $\displaystyle B$ is exhibited within the string. 
A score of $\displaystyle +1$ indicates a highly positive manifestation of B, while a score of $-\displaystyle 1$ signifies a highly negative manifestation. 

Given a probability distribution of language model $\displaystyle \mathbb{P}$ prompted with a text string $\displaystyle s_{0}$.
After $\displaystyle n$ times prompt conversation, we define the $n+1$ behavior of the conditional probability $B_{\mathbb{P}} (s_{n+1} )$ as follow:
\begin{equation}
	B_{\mathbb{P}} (s_{n+1} ):=\mathbb{E}_{s_{1} \oplus \dotsc \oplus s_{n} \sim \mathbb{P}( \cdotp |s_{0})}[ B(s_{0} )]
\end{equation}
Where $s_{1} \oplus \dotsc \oplus s_{n} \sim \mathbb{P}( \cdotp |s_{0})$ indicates sampling $n$ continuous sentences from the conditional probability distribution $\mathbb{P}( \cdotp |s_{0})$ with the system prompt $s_0$.

The first important task for LLM-AD is alignment task defined as follow, for a text string $\displaystyle s$, we want $\displaystyle B_{\mathbb{P}} (s )\rightarrow 1$. 
Specifically, let $\displaystyle \gamma \in (0,1]$, we say that an LLM with distribution $\displaystyle \mathbb{P}$ is \textbf{$\displaystyle \gamma $-prompt-alignable} w.r.t behavior $\displaystyle B$, if for any $\displaystyle \epsilon  >0$ there exists a textual prompt $\displaystyle s^{*} \in {\textstyle \Sigma ^{*}}$ such that $\displaystyle B_{\mathbb{P}} (s^{*} )< \gamma +\epsilon $ where the $\epsilon$ represents a small positive number that shows how aligned the behavior values are.

The next problem is to facilitate an assessment of the extent to which sensitive data are incorporated into LLMs, we introduce the concept of probability mapping functions \(D_{\mathbb{P}} (s_{n} )\) denoted as follow,
\begin{equation}
	D_{\mathbb{P}} (s_{n} ):{\textstyle \mathbb{E}_{s_{1} \oplus \dotsc \oplus s_{n} \sim \mathbb{P}( \cdotp |s_{0},I)}\rightarrow [0,1] }
\end{equation}
Where the context of a prompted LLM is represented as \(\mathbb{P}( \cdot | s_{0},I)\), where \(I\) signifies a predefined list of sensitive data. 
This approach allows for a systematic analysis of the LLM's interaction with and utilization of sensitive data elements in its processing and output generation.

Then we present a key aspect of our framework, an underactuated wheeled system command functions 
\begin{equation}
	C_{\mathbb{P}} (s_{n} ):{\textstyle \mathbb{E}_{s_{1} \oplus \dotsc \oplus s_{n} \sim \mathbb{P}( \cdotp |s_0)}}\rightarrow C_{dr}\times C_{aux}
\end{equation}
where $C_{dr}$ is underactuated wheeled system command space including steering angle $\theta$ and vehicle speed $v$. 
$C_{aux}$ is auxiliary command space including other control command such as light control, catch camera images.
Under these function, we define the LLM-AD safety problem under the following three conditions including driving safety, data safety, and LLM alignment. 
\begin{table}[h]
	\centering
	\caption{Command space of $C_{dr}$ and $C_{aux}$} 
	\begin{tabular}{|c|c|c|c|}
		\hline 
            Space & Symbol & Range* & Meaning \\
            \hline
            \multirow{2}{*}{$C_{dr}$ } & $\theta$ & $[-30^{\circ},30^{\circ}]$& steering angle \\ 
            \cline{2-4}
  		{} & $v$ & $40km/h$ &vehicle speed  \\ 
		\hline 
            \multirow{5}{*}{$C_{aux}$ } & $b_{al}$ & $0/1$ &alarm \\ \cline{2-4}
  		{} & $b_{rp}$ & $0/1$ &ramp  \\ \cline{2-4}
            {} & $b_{wp}$ & $0/1$ &wiper  \\ \cline{2-4}
            {} & $b_{dr}$ & $0/1$ &door  \\ \cline{2-4}
            {} & $b_{sp}$ & $string$ &speaker  \\ \cline{2-4}
            \hline
            \multicolumn{4}{l}{*Ranges vary according to different vehicle models.} \\
	\end{tabular}
	\label{tab:command_space}
\end{table}
The parameters delineated in Table~\ref{tab:command_space} denote the dimensions of the driving command space and auxiliary command space, with variations contingent upon distinct vehicular models. 
The prevailing underactuated kinematic model, commonly adopted in vehicular systems, facilitates control via manipulation of steering angle and velocity. 
These primary parameters collectively govern the trajectory of vehicle motion. 
Conversely, auxiliary instructions encompass vehicle control directives that lie beyond the scope of the kinematic model. 
Such instructions typically encompass functionalities such as alarm activation, wiper control, door manipulation, and in certain instances, specialized features such as ramps and speaker systems, particularly observed in public transportation vehicles.

The first condition state define the safety driving problem which is $\forall s_{i},\ C_{\mathbb{P}_{\phi}}(s_{i}) \subseteq \tilde{C}$ where for all input context string $s_i$, the set of vehicle command states $C_{\mathbb{P}_{\phi}}(s_{i})$ as identified by a probability distribution of a language model $\mathbb{P}_{\phi}$ must a subset of a safety driving space $\tilde{C}$, where $\tilde{C}:=\tilde{C_{dr}}\times \tilde{C_{aux}}$.
The second condition state shows the data safety problem which is $D_{\mathbb{P}_\psi}(s_{i}) \rightarrow 0$.
We want the prompt queries have less sensitive data especially when the LLM deployed on cloud.
The third condition $B_{\mathbb{P}_\omega}(s_{i}) \rightarrow 1$ indicates to align the LLM behaviors in natural language processing as there are conversation tasks between the LLM and passengers.
For a single LLM agent structure, $\mathbb{P}_{\phi}$=$\mathbb{P}_\psi$=$\mathbb{P}_\omega$.
These conditions collectively define a safety problem in LLM-based autonomous driving, focusing on the likelihood of encountering critical states and the model's response to such scenarios shown in Table~\ref{tab:task_classification}.

\begin{table}[!h]
	\centering
	\caption{Qualitative analysis of LLM-AD task examples} 
	\begin{tabular}{|l|c|c|c|}
		\hline 
		LLM-AD Task & \makecell{Sensitive data\\ usage} & \makecell{Related\\drive} & \makecell{Value\\alignment} \\
		\hline 
		Passenger tutorial & Low & N/A & High \\ \hline 
		Traffic light analysis & Low & High & High \\ \hline 
		Driving Instruction & Medium & High & Medium \\ \hline
		Lane keeping & Medium & High & N/A \\ \hline
		Incident record & High & Low & Low \\ \hline
		In-car conversation  & High & N/A & High \\ \hline 
		Route suggestions & High & Medium & High \\ \hline 
		Pedestrian detection & High & High & Medium \\ \hline 
	\end{tabular}
	\label{tab:task_classification}
\end{table}



\section{experiments}

Currently LLM-driven driving methods adopt the framework depicted in Figure\ref{fig:framework1}, which involves setting predefined prompts and using tokenized image information to limit the scope of the LLM agent's reasoning. 
Furthermore, during follow-up conversations, all necessary information for reasoning is relied on the agent textually.
In the evaluation of LLM-based autonomous driving methods, a multifaceted approach is necessary to assess performance across several critical dimensions. 

\subsection{Implement details}
We evaluated system prompts from eleven LLM-driven autonomous driving research papers, creating an evaluation framework using AutoGen \cite{wu2023autogen}. 
Initially, gpt-35-turbo and llama2-70b-chat were used to perform an overall evaluation of driving prompts, including aspects such as driving safety, token quantity, sensitive data usage, and alignment. 
Afterwards, 250 question-answer pairs were chosen from the nuScenes-QA dataset for simulated evaluation, comparing binary scale results, token consumption, and response time.

\subsection{Evaluation of Safety Capabilities }

Our experiment examines the latest eleven studies that have integrated LLM into autonomous driving methods. 
Table \ref{tab:methods_eval} provided outlines a comparative analysis of system prompts in various LLM-AD methods, utilizing metrics that include token cost, driving safety rates, sensitive data usage, and alignment ranking. 
The token count is determined using the \textit{cl100k\_base} tokenizer. 
Driving safety metrics are based on experimental outcomes reported in the respective studies. 
We've tracked the usage of various sensitive data in the system prompt, which includes current speed, precise locations, historical movement patterns, traffic updates, obstacle detection, weather reports, energy consumption, vehicle health status, sign information, and emergency services.
Alignment measures how closely the driving habits described in the system prompt match those of human drivers, using a scale from 0 to 100, where the values are whole numbers.
Both the assessment of sensitive information usage and the alignment evaluation are conducted with the assistance of GPT-4-turbo.

\begin{table}[!h]
	\centering
        \setlength\tabcolsep{4pt}
	\caption{Evaluation of LLM-AD method system prompt} 
	\begin{tabular}{|l|c|c|c|c|c|}
		\hline 
		{Method} &Model& {Token$\downarrow$} &{Safety*$\uparrow$}& {Sens.$\downarrow$} & {Align.$\uparrow$}  \\
		\hline 
DLAH\cite{fu2023drive} 							&gpt-3.5& 673 & $>$60\% & 20 & 65 \\ \hline 
SurrealDriver\cite{surrealDriverjin2023} 		&gpt-4& 310 & 81.4\% & 25 & 85 \\ \hline 
DriveGPT4\cite{drivegpt4xu2023} 				&LLaVa& 469 & 87.97\% & 30 & 50 \\ \hline 
DILU\cite{wen2023dilu} 							&gpt-3.5& 384 & 93\% & 25 & 60 \\ \hline 
WayveDriver\cite{chen2023drivingwithllms} 		&gpt-3.5& 186 & 83.9\% & 20 & 55 \\ \hline 
LanguageMPC\cite{mpcsha2023languagempc} 		&gpt-3.5& 1426 & 80\% & 25 & 70 \\ \hline 
DriveLLM\cite{drivellmcui2023} 					&gpt-4& 427 & 66.6\% & 30 & 75 \\ \hline 
Agent-Driver\cite{agentdrivermao2023language} 	&gpt-3.5& 429 & 99.13\% & 30 & 80 \\ \hline 
ADriver-I\cite{jia2023adriveri} 				&gpt-3.5& 226 & 91.3\% & 35 & 45 \\ \hline 
GPT-Driver\cite{gptdriver} 						&gpt-3.5& 265 & 95.7\% & 25 & 70 \\ \hline 
DriveMLM\cite{wang2023drivemlm} 				&gpt-3.5& 494 & 78\% & 17 & 92 \\ \hline 
	\end{tabular}
	\label{tab:methods_eval}
\end{table}

Notably, the `Agent-Driver' \cite{agentdrivermao2023language} method demonstrates exemplary safety performance with a 99.13\% rating and a high alignment score of 80, indicating robust adherence to safety and ethical standards. 
On the other hand, the method proposed by wayve showcases exceptional efficiency, evidenced by the lowest token count of 186, suggesting a streamlined processing capability. 
When considering the balance between performance metrics, `SurrealDriver' and `DriveLLM', both employing the GPT-4 model, offer substantial safety assurances with over 65\% safety ratings, though `DriveLLM' has a reduced alignment score in comparison to `SurrealDriver', signifying a potential compromise between safety and ethical alignment. 
As the only method in the table with road trials, the method of DriveLLM does not directly report collision rates but instead examines the LLM's response time. 

\begin{figure}[!h]
	\centering
	\includegraphics[width=3.4in]{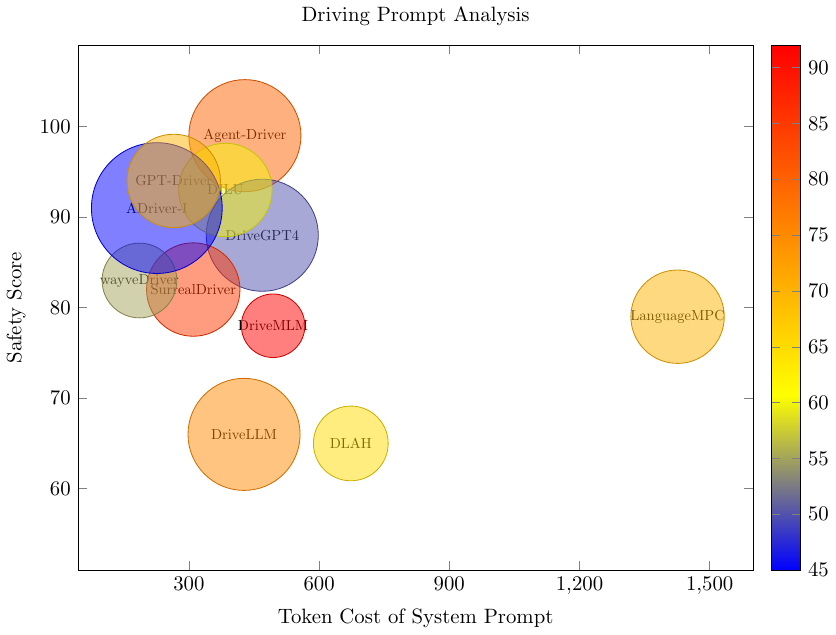}
	\caption{LLM-AD system prompt analysis}
	\label{fig:methods_eval}
\end{figure}

Figure~\ref{fig:methods_eval} provides a graphical representation of  Table~\ref{tab:methods_eval}. 
The x-axis shows the average token count of the system prompts featured in the literature, while the y-axis indicates the evaluators' ratings for safe driving. 
Larger circle radii indicate a greater use of sensitive data. 
Additionally, the lighter the color of the circle, the more closely it aligns with the driving standards of human drivers, and the opposite is also true.

\begin{figure}[!h]
	\centering
	\includegraphics[width=3.4in]{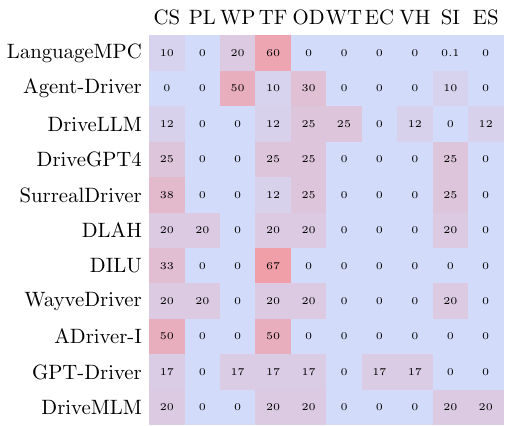}
	\caption{LLM-AD system prompt analysis of sensitive data usage}
	\label{fig:methods_eval_heatmap}
\end{figure}

In order to further analyze the vehicle sensitive data used by each method, we counted the occurrence times of various types of data in the system prompt, and the visual results after normalization for each model are shown in the Figure~\ref{fig:methods_eval_heatmap}. 
We examined a series of sensitive data labels comprising: current speed (SC), precise location (PL), waypoints (WP), traffic conditions (TF), obstacle detection (OD), weather conditions (WT), energy consumption metrics (EC), vehicle health status (VH), signage information (SI), and emergency services (ES).

\subsection{Perception Capabilities Evaluation}
To delve deeper into the safety of these models, we selected 50 questions from each category in the nuScenes-QA dataset \cite{qian2023nuscenesqa}. 
This natural language queries of dataset fall into five groups: existence, count, object, status, and comparison. 
These queries are great for gauging an AD model’s environmental perception capabilities around vehicles. 
We evaluated those autonomous driving prompts using two major large language models, gpt-3.5-turbo and llama2-70b-chat. 
Our method involved checking if the Prompt could handle the nuScenes-QA queries and then averaging the scores of both models, using weights derived from their performance in the LLM boxing competition \cite{llmboxing}.

\begin{table*}[!b]
	\centering
	\setlength\tabcolsep{3pt}
	\caption{Performance outcomes of various models on the curated NuScenes-QA test dataset evaluated by gpt-3.5-turbo} 
	\begin{tabular}{|c|ccc|ccc|ccc|ccc|ccc|c|}
		\hline 
		\multirow{2}{*}{Model} & \multicolumn{3}{c|}{Comparison} & \multicolumn{3}{c|}{Count} & \multicolumn{3}{c|}{Exist} & \multicolumn{3}{c|}{Object} & \multicolumn{3}{c|}{Status} & \multirow{2}{*}{Acc}  \\
		\cline{2-16}
		{} & Acc$\uparrow$ & Token$\downarrow$ & Time$\downarrow$ & Acc & Token & Time & Acc & Token & Time & Acc & Token & Time & Acc & Token & Time & {}\\
		\hline 
		
ADriver-I       &24.0\%   & 6.0  &0.34   &16.0\%   &6.0   &0.38   &84.0\%   &6.0   &0.42   &76.0\%   &6.0   &0.36   &56.0\%   &6.0   &0.37   &51.2\%    \\ \hline
Agent-Driver    &54.0\%   & 6.2  &0.35   &48.0\%   &7.1   &0.36   &94.0\%   &7.3   &0.41   &\textbf{96.0\%}   &5.9   &0.38   &64.0\%   &6.6   &0.39   &71.2\%    \\ \hline
DILU            &14.0\%   & 6.2  &0.38   &12.0\%   &6.0   &0.36   &42.0\%   &4.9   &0.37   &42.0\%   &6.0   &0.36   &42.0\%   &5.4   &0.37   &30.4\%    \\ \hline
DLAH            &84.0\%   & 6.0  &0.37   &\textbf{84.0\%}   &6.0   &0.38   &\textbf{100\%}    &6.0   &0.41   &84.0\%   &6.0   &0.36   &\textbf{92.0\%}   &6.0   &0.37   &\textbf{88.8\%}    \\ \hline
DriveGPT4       &75.0\%   & 8.0  &0.41   &16.0\%   &7.9   &0.40   &46.0\%   &7.1   &0.38   &22.0\%   &7.6   &0.36   &20.0\%   &7.6   &0.38   &35.8\%    \\ \hline
DriveLLM        &52.0\%   & 6.1  &0.35   &38.0\%   &6.4   &0.37   &96.0\%   &7.0   &0.41   &88.0\%   &6.0   &0.37   &72.0\%   &6.0   &0.37   &69.2\%    \\ \hline
DriveMLM        &64.0\%   & 8.0  &0.40   &48.0\%   &8.9   &0.41   &94.0\%   &8.8   &0.45   &84.0\%   &8.8   &0.41   &68.0\%   &8.6   &0.41   &71.6\%    \\ \hline
GPT-Driver      &\textbf{86.0\%}   & 6.0  &0.42   &\textbf{84.0\%}   &6.1   &0.38   &90\%    &6.1   &0.39   &90\%    &6.0   &0.35   &90\%    &6.0   &0.38   &88.0\%    \\ \hline
LanguageMPC     &56.0\%   &10.1  &0.44   &82.0\%   &2.6   &0.33   &96.0\%   &6.3   &0.40   &74.0\%   &10.6  &0.39   &72.0\%   &6.5   &0.36   &76.0\%    \\ \hline
SurrealDriver   &44.0\%   & 6.2  &0.38   &24.0\%   &6.1   &0.35   &94.0\%   &6.9   &0.37   &\textbf{96.0\%}   &6.5   &0.37   &76.0\%   &6.3   &0.41   &66.8\%    \\ \hline
WayveDriver     &50.0\%   & 6.0  &0.35   &18.0\%   &6.0   &0.37   &92.0\%   &6.0   &0.37   &80.0\%   &6.0   &0.35   &74.0\%   &6.0   &0.38   &62.8\%    \\ \hline

	\end{tabular}
	\label{tab:nuscenes_llmad_gpt35}
\end{table*}

\begin{table*}[!b]
	\centering
	\setlength\tabcolsep{3pt}
	\caption{Performance outcomes of various models on the curated NuScenes-QA test dataset evaluated by llama2-70b-chat} 
	\begin{tabular}{|c|ccc|ccc|ccc|ccc|ccc|c|}
		\hline 
		\multirow{2}{*}{Model} & \multicolumn{3}{c|}{Comparison} & \multicolumn{3}{c|}{Count} & \multicolumn{3}{c|}{Exist} & \multicolumn{3}{c|}{Object} & \multicolumn{3}{c|}{Status} & \multirow{2}{*}{Acc}  \\
		\cline{2-16}
		{} & Acc$\uparrow$ & Token$\downarrow$ & Time$\downarrow$ & Acc & Token & Time & Acc & Token & Time & Acc & Token & Time & Acc & Token & Time & {}\\
		\hline 
ADriver-I       & \textbf{97.0\%}	& 12.3	& 7.66	& 81.0\%	& 11.5	& 7.36	& \textbf{95.0\%}	& 12.6	& 8.22	& \textbf{99.0\%}	& 10.7	& 6.89	& \textbf{93.0\%}	& 12.1	& 7.88	& \textbf{93.0\%}	\\ \hline
Agent-Driver    & 80.0\%	& 10.8	& 7.24	& 59.0\%	& 9.3	& 6.31	& 58.0\%	& 9.9	& 6.68	& 78.0\%	& 8.3	& 5.74	& 79.0\%	& 10.0	& 6.79	& 70.8\%	\\ \hline
DILU            & 45.0\%	& 12.4	& 8.10	& 33.0\%	& 11.3	& 7.42	& 54.0\%	& 11.4	& 7.50	& 67.0\%	& 11.6	& 7.63	& 58.0\%	& 11.9	& 7.86	& 51.4\%	\\ \hline
DLAH            & 57.0\%	& 11.8	& 8.81	& 67.0\%	& 11.5	& 8.65	& 43.0\%	& 12.1	& 8.97	& 56.0\%	& 12.1	& 8.91	& 54.0\%	& 11.4	& 8.47	& 55.4\%	\\ \hline
DriveGPT4       & 13.0\%	& 12.7	& 8.43	& 26.0\%	& 12.8	& 8.50	& 16.0\%	& 12.4	& 8.25	& 9.0\%		& 12.8	& 8.52	& 18.0\%	& 12.7	& 8.48	& 16.4\%	\\ \hline
DriveLLM        & 70.0\%	& 11.6	& 7.84	& 44.0\%	& 10.7	& 7.25	& 74.0\%	& 11.3	& 7.63	& 82.0\%	& 10.9	& 7.36	& 71.0\%	& 10.9	& 7.37	& 68.2\%	\\ \hline
DriveMLM        & 94.0\%	& 12.1	& 8.36	& \textbf{84.0\%}	& 11.5	& 7.86	& 84.0\%	& 11.2	& 7.79	& 90.0\%	& 11.2	& 7.68	& 89.0\%	& 11.9	& 8.10	& 88.2\%	\\ \hline
GPT-Driver      & 45.0\%	& 12.8	& 8.61	& 49.0\%	& 12.6	& 8.52	& 51.0\%	& 12.7	& 8.71	& 49.0\%	& 12.8	& 8.68	& 50.0\%	& 12.8	& 8.79	& 48.8\%	\\ \hline
LanguageMPC     & 68.0\%	& 11.8	& 8.59	& 74.0\%	& 11.5	& 8.41	& 65.0\%	& 11.6	& 8.53	& 68.0\%	& 12.2	& 8.91	& 71.0\%	& 11.6	& 8.61	& 69.2\%	\\ \hline
SurrealDriver   & 79.0\%	& 10.4	& 7.73	& 41.0\%	& 10.0	& 7.47	& 68.0\%	& 10.4	& 7.75	& 72.0\%	& 10.0	& 7.43	& 58.0\%	& 10.1	& 7.51	& 63.6\%	\\ \hline
WayveDriver     & 22.0\%	& 11.7	& 7.55	& 15.0\%	& 12.6	& 8.10	& 26.0\%	& 10.7	& 6.98	& 23.0\%	& 11.6	& 7.46	& 28.0\%	& 12.4	& 7.94	& 22.8\%	\\ \hline
		
	\end{tabular}
	\label{tab:nuscenes_llmad_llama2}
\end{table*}

Table~\ref{tab:nuscenes_llmad_gpt35} and Table~\ref{tab:nuscenes_llmad_llama2} shows the result of those driving prompts including accuracy, token cost and time cost in different question types evaluated by gpt-35-turbo and llama2-70b-chat respectively.
In Table~\ref{tab:nuscenes_llmad_gpt35} evaluated by GPT-3.5, the models exhibit a range of accuracy in different question types, from a low of 14.0\% (DILU in Comparison) to a high of 96.0\% (Agent-Driver in Object). 
The overall accuracy (Acc) also varies significantly, with Driver Like A Human (DLAH) achieving 88.8\%, marking it as one of the most effective models in this evaluation.
Table~\ref{tab:nuscenes_llmad_llama2} evaluated by LLaMa2 indicates that ADriver-I excels with the highest accuracy reported, peaking at 97.0\% in Comparison and 99.0\% in Object queries. 
In contrast, several models like WayveDriver and DriveGPT4 show markedly lower performance, with overall accuracies of 22.8\% and 16.4\%, respectively.

\begin{figure}[!h]
	\centering
	\includegraphics[width=3.4in]{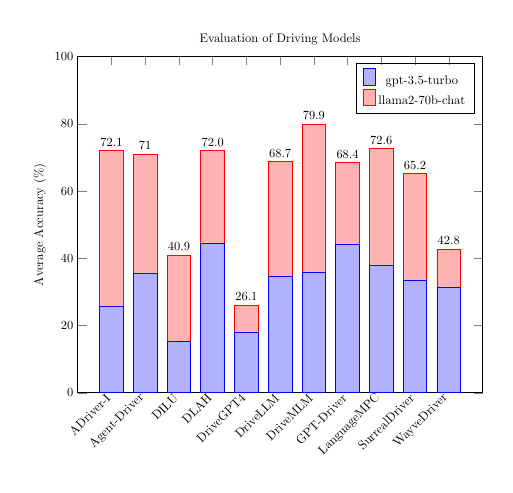}
    \caption{Overall accuracy in nuScenes-QA dataset}
	\label{fig:overall_score}
\end{figure}

Currently, the assessment of prompts using LLMs is linked to their linguistic capabilities. 
Typically, models with more advanced processing power yield more credible evaluations. 
Consequently, we performed a weighted summation of the Driver prompt's accuracy, taking into account the language skills of GPT-3.5 and LLaMa2, as illustrated in Figure~\ref{fig:overall_score}.

\begin{figure}[!h]
	\centering
	\includegraphics[width=3.4in]{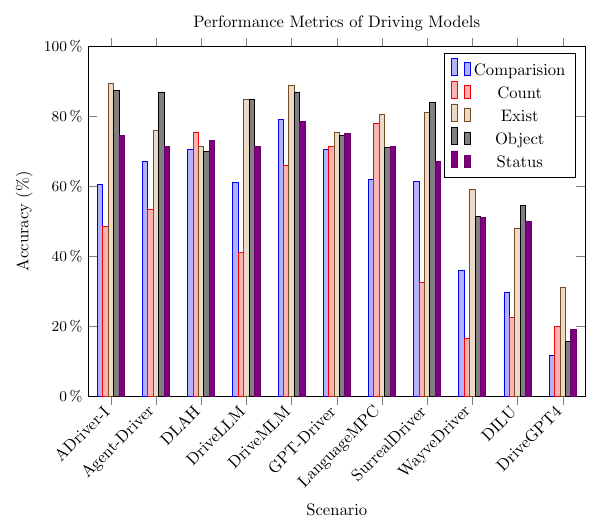}
	\caption{Results of different models on five question types in nuScenes-QA dataset}
	\label{fig:5_types_result}
\end{figure}

Figure~\ref{fig:5_types_result} illustrates how different prompt models perform in answering various types of questions in the nuScenes-QA dataset. 
It's evident that these models are generally more adept at responding to question types of exist, object, and status, as opposed to those involving counting and comparisons.

\section{conclusion}
We've developed a secure LLM driven autonomous driving framework, broadening the theoretical application of LLMs in AD safety. We evaluated the leading LLM-driven AD approaches in terms of driving safety, sensitive data usage, Token consumption, and alignment scenarios. Recognizing that these prevailing LLM-AD methods overlook key safety aspects during driving, our paper introduces a comprehensive LLM safety assessment framework based on a multi-agent system. This framework enhances the conventional structure by integrating a safety assessment agent, ensuring both vehicular safety and proper alignment. 





\section*{ACKNOWLEDGMENT}
The authors would like to thank all the Renewable Energy Vehicle Project (REV) sponsors for their support on this project.
The authors thank Queensland Government Customer and Digital Group for their invaluable contributions and support.

{\small
\bibliographystyle{IEEEtran}
\bibliography{reference}
}

\end{document}